\documentclass[letterpaper]{article} 
\usepackage{aaai2026}  
\usepackage{times}  
\usepackage{helvet}  
\usepackage{courier}  
\usepackage[hyphens]{url}  
\usepackage{graphicx} 
\usepackage{natbib}
\usepackage{caption} 
\usepackage{algorithm}
\usepackage{booktabs}
\usepackage[table]{xcolor}
\usepackage{makecell}
\usepackage{algpseudocode}
\usepackage{amsmath}

\usepackage{multirow}
\usepackage{siunitx}
\usepackage{amssymb}
\usepackage{newfloat}
\usepackage{listings}
\DeclareCaptionStyle{ruled}{labelfont=normalfont,labelsep=colon,strut=off} 
\floatstyle{ruled}
\newfloat{listing}{tb}{lst}{}
\floatname{listing}{Listing}
\title{Leveraging Failed Samples: A Few-Shot and Training-Free Framework for Generalized Deepfake Detection}
\author{
    Shibo Yao\textsuperscript{\rm 1}, 
    Renshuai Tao\textsuperscript{\rm 1}\thanks{Corresponding author.}, Xiaolong Zheng\textsuperscript{\rm 2}, Chao Liang\textsuperscript{\rm 3}, Chunjie Zhang\textsuperscript{\rm 1}
}
\affiliations{

    \textsuperscript{\rm 1}Institute of Information Science, Beijing Jiaotong University\\
    \textsuperscript{\rm 2}Institute of Automation, Chinese Academy of Sciences\\
    \textsuperscript{\rm 3}School of Computer Science, Wuhan University\\
}

\usepackage{bibentry}

\begin{document}

\maketitle

\begin{abstract}
Recent deepfake detection studies often treat unseen sample detection as a ``zero-shot" task, training on images generated by known models but generalizing to unknown ones. A key real-world challenge arises when a model performs poorly on unknown samples, yet these samples remain available for analysis. This highlights that it should be approached as a ``few-shot" task, where effectively utilizing a small number of samples can lead to significant improvement. Unlike typical few-shot tasks focused on semantic understanding, deepfake detection prioritizes image realism, which closely mirrors real-world distributions. In this work, we propose the Few-shot Training-free Network (FTNet) for real-world few-shot deepfake detection. \textbf{Simple yet effective}, FTNet differs from traditional methods that rely on large-scale known data for training. Instead, FTNet uses \textbf{only one fake sample} from an evaluation set, mimicking the scenario where new samples emerge in the real world and can be gathered for use, without any training or parameter updates. During evaluation, each test sample is compared to the known fake and real samples, and it is classified based on the category of the nearest sample. We conduct a comprehensive analysis of \textbf{AI-generated images from 29 different generative models} and achieve a new SoTA performance, with an average \textbf{improvement of 8.7\%} compared to existing methods. This work introduces a fresh perspective on real-world deepfake detection: when the model struggles to generalize on a few-shot sample, leveraging the failed samples leads to better performance.\footnote{The code will be open-sourced upon publication.}
 

\end{abstract}


\section{Introduction}
With the rapid development of deep learning-based generative techniques, especially the rise of advanced models such as Generative Adversarial Networks (GANs) \cite{goodfellow2014generative,  rossler2019faceforensics++, choi2018stargan} and Diffusion models \cite{ rombach2022high, nichol2021improved, podell2023sdxl, esser2024scaling}, AI-generated images have made significant strides in producing realistic and diverse content. These models, which have evolved over the years, are capable of generating highly convincing synthetic media, including faces and entire video sequences that are nearly indistinguishable from real content.

However, AI-generated content, particularly deepfakes, has become a significant threat to information integrity, public trust, and digital security. Deepfake technology allows for the creation of hyper-realistic fake media, which can be easily shared and distributed, making it difficult for the public to distinguish between what is real and what is fabricated. The implications are far-reaching: deepfakes can be used for spreading false propaganda, manipulating public opinion, committing commercial fraud, or even defaming individuals by altering their appearance or words in a video.

\begin{figure}[!t]
  \centering
   \includegraphics[width=\columnwidth]{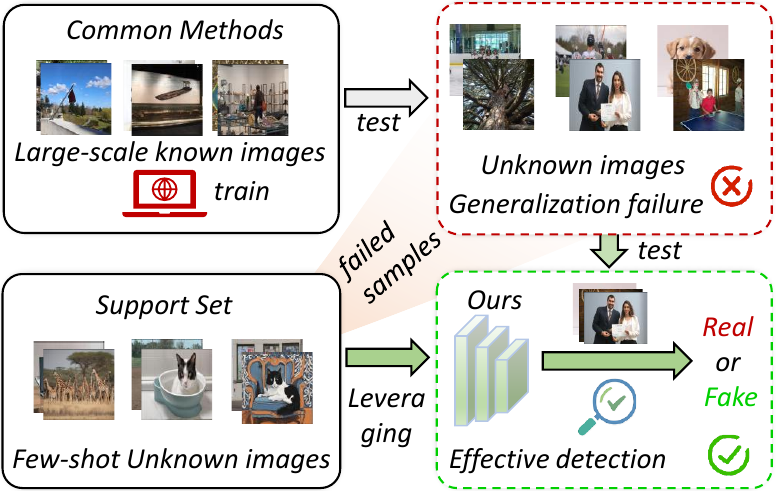}
   \caption{Comparison of traditional methods and our few-shot method in real-world scenarios.}  
   \vspace{-0.2in}
   \label{fig:fig1}
\end{figure}
Moreover, as the quality of generative models continues to improve, detecting deepfakes has become increasingly challenging. Traditional detection methods\cite{tan2024rethinking, yan2024orthogonal, guillaro2025bias,tao2025sagnet} that rely on identifying artifacts or inconsistencies in synthetic images and videos are gradually losing effectiveness as generative models become more sophisticated. For instance, deepfake videos created with state-of-the-art GANs or Diffusion models often exhibit a high level of realism, making it harder to distinguish them from real media. This creates a pressing need for deepfake detection systems that are not only capable of detecting known types of deepfakes but also generalize well to new, unseen variants. In response to this challenge, deepfake detection research has focused on developing methods that can adapt to novel deepfake samples. Many existing techniques treat the detection of unseen samples as a ``zero-shot" task, in which models are trained on a set of known deepfakes and expected to generalize to unknown ones. However, this approach often falls short in real-world scenarios, where new deepfakes are continuously created, and the model’s performance degrades when exposed to previously unseen content. This highlights the importance of developing more robust detection methods that can handle unseen deepfakes effectively, without relying on extensive retraining or large-scale datasets. As shown in Figure \ref{fig:fig2}(a), the baseline detector overfits to the known generative model and has difficulty distinguishing unknown generative models. Instead, our method (Figure \ref{fig:fig2}(b)) classifies various generative models well, which indicates its great potential in addressing the above challenges.


In this work, we introduce the Few-shot Training-free Network (FTNet), a novel approach for real-world deepfake detection. One promising direction in deepfake detection is the use of ``few-shot" learning approaches, where the model is trained to recognize deepfakes from only a small number of examples. Few-shot learning can be particularly useful in real-world scenarios, where \textbf{new deepfake samples may emerge and can be quickly gathered for analysis}. Unlike traditional methods that require training on large amounts of labeled data, few-shot detection techniques can adapt to new deepfake types with minimal resources. However, deepfake detection using few-shot learning presents unique challenges, primarily because \textbf{deepfakes focus on mimicking the realism of images, which closely resemble real-world distributions}. This makes it difficult for conventional few-shot learning methods, which often focus on semantic understanding, to perform effectively in deepfake detection. 

The proposed FTNet addresses these challenges by operating in a training-free manner. It uses only a minimal number of samples, specifically \textbf{one fake sample} from the evaluation set, mimicking real-world scenarios where new deepfakes emerge and can be quickly gathered for use. The features and labels of these samples are then injected into a dynamic knowledge base, which we call the key-value cache. This approach eliminates the need for parameter updates or retraining, significantly reducing computational overhead and resource requirements. During the evaluation phase, each test sample is compared against the known fake and real samples and is classified based on the category of the nearest sample. FTNet’s simple yet effective method allows it to rapidly adapt to new deepfake samples, offering a scalable and computationally efficient solution for detecting.


Through extensive experiments on \textbf{AI-generated images from 29 different generative models}, we demonstrate that the proposed FTNet achieves new state-of-the-art performance in deepfake detection, outperforming existing methods by \textbf{8.7\%} (with the fine-tuning version, FTNet-T, achieving 12.1\%). This work highlights the potential of using a small-sample, training-free method to enhance deepfake detection, especially in scenarios where new deepfakes are constantly being generated. By focusing on effectively leveraging few-shot samples, FTNet provides a scalable and practical solution to the growing problem of deepfake detection. The main contributions are summarized as follows:

\begin{figure}[!t]
  \centering
   \includegraphics[scale=0.32]{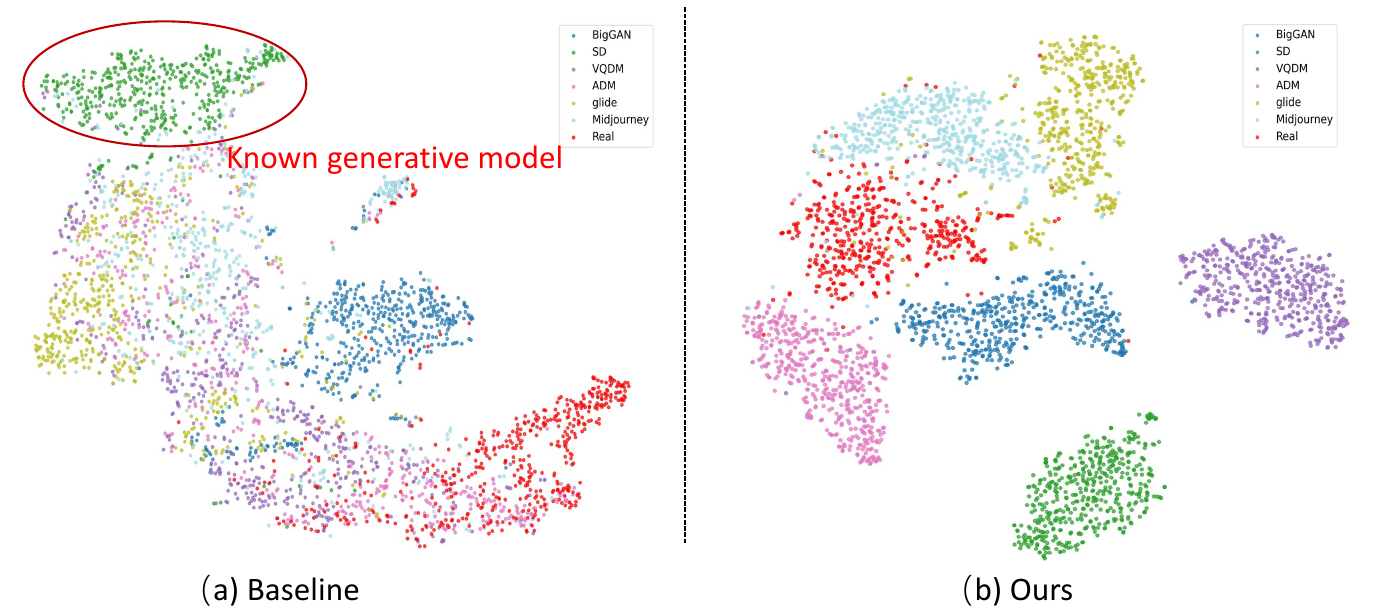}
   \caption{Comparison of traditional detector and our method on a 6-class generative model feature space.}  
   \label{fig:fig2}
\end{figure}

\begin{itemize}
\item We propose the FTNet, a novel approach for deepfake detection that uses a minimal number of samples (only one fake and one real sample) without any need for retraining or parameter updates. This approach significantly reduces the need for large-scale data and training resources.
\item FTNet is designed to handle real-world deepfake detection, where new, unseen deepfake samples frequently emerge. By utilizing a few-shot, training-free methodology, FTNet adapts to new deepfake types effectively, providing a practical solution in dynamic environments.
\item Through extensive experiments on AI-generated images from 29 different generative models across three open-world datasets, demonstrating the effectiveness.
\end{itemize}


\begin{figure*}[!t]
\centering
\includegraphics[width=\textwidth]{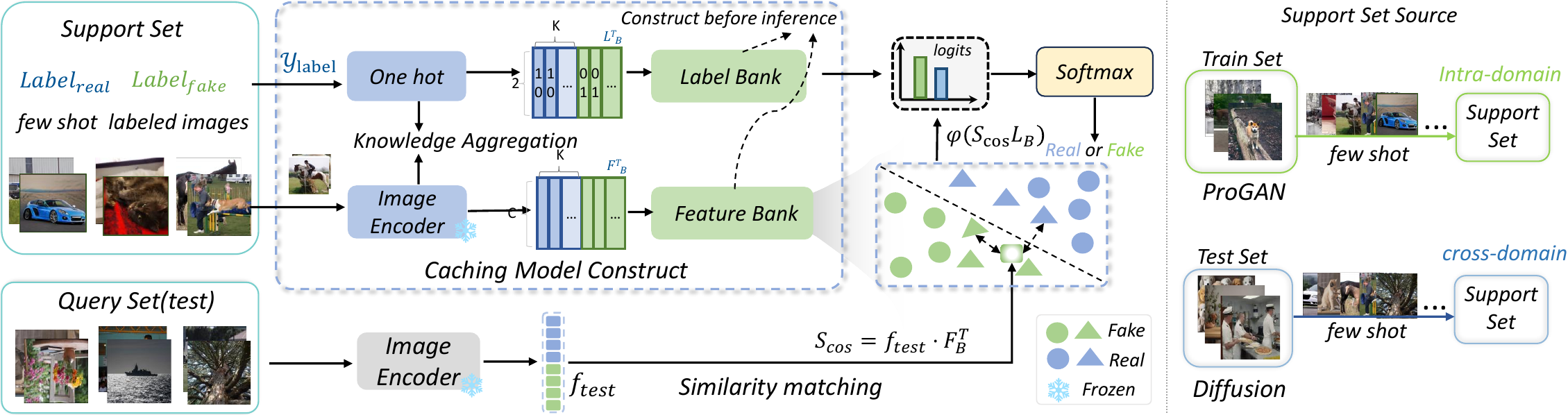} 
\vspace{-0.1in}
\caption{Overall framework of our proposed FTNet. The core of FTNet extracts features from the intermediate layers of the CLIP image encoder to build a cache module, leveraging few-shot labeled samples for efficient detection of unknown samples.}
\vspace{-0.1in}
\label{fig3}
\end{figure*}

\section{Related Work}
Owing to space constraints in the main document, this section has been included in the supplementary materials.

\section{Method}
In this section, we reframe the problem as a few-shot detection challenge. We will present its core components in order: the Cache Model Construction, our training-free FTNet, and its fine-tuned variant FTNet-T.
\subsection{Problem Definition}
The core challenge faced by the deepfake detection field is that existing methods usually rely on large-scale data training of a single generative model (such as ProGAN\cite{karras2017progressive} or a specific diffusion model\cite{ho2020denoising}), which quickly becomes outdated in the context of the diversification and rapid evolution of AI generation technology. With the continuous emergence of new generative models (such as upgraded GAN variants, iterative versions of diffusion models, etc.), the forgery traces of their synthetic images are more hidden and show unique artifact characteristics, resulting in the serious lack of generalization ability of traditional methods due to the large domain difference between the source domain and the target domain. In addition, in real scenarios, we can often obtain a small number of samples of new generative models, while existing methods either require a large amount of labeled data for training or cannot use limited samples to update detection capabilities due to the lack of dynamic adaptive mechanisms. This reliance on large-scale homogeneous training data is in sharp conflict with the actual needs of sample scarcity and continuous evolution of generative models in the real world. By reconceptualizing deepfake detection as a few-shot classification task, that is, treating images from different generators as different categories, we can simplify this complex problem. This approach allows us to extract rich features from very few samples and achieve automatic adaptation to unseen biases without further large-scale fine-tuning or large datasets. Therefore, there is an urgent need for a detection framework that can achieve fast adaptation with a small number of target domain samples without further training. This problem definition lays the foundation for subsequent few-shot detection methods.

\subsection{Framework Overview}
To address the challenge of the lack of generalization ability of existing deepfake detections when facing unknown new models, we propose a novel few-shot detection framework. As shown in Figure \ref{fig3}. This framework uses the pre-trained CLIP model as the visual backbone, builds an efficient cache model, and uses a small number of target domain samples to identify the synthetic images of the newly generated model. It is suitable for open-world scenarios with scarce data.
We first propose FTNet without parameter training, which identifies newly generated images by building a key-value cache. It is suitable for scenarios with extremely limited resources, real-time deployment, or extremely scarce samples. On this basis, we propose FTNet-T, which breaks through the performance ceiling by fine-tuning the cache model at low cost. It is suitable for scenarios that pursue high precision, have sufficient samples, or can be updated in non-real time.
FTNet and FTNet-T form a complete paradigm of ``adaptive knowledge injection" : FTNet lays the foundation for instant caching of new fake knowledge into the CLIP feature space, and FTNet-T optimizes knowledge through lightweight fine-tuning to achieve a seamless transition from basic adaptation to precise detection, ensuring the efficiency of the framework in a variety of practical scenarios.

\subsection{Cache Model Construction}
Past few-shot deepfake detection method\cite{wu2025few} treated different generative models as different categories, and unified real images into one category. However, this setting results in them only being able to test a single dataset during the test, which is not in line with real scenarios. Real-world detection scenarios are often characterized by a mixture of images from multiple, diverse, and often unknown sources. In our task, although we treat the generative model as an independent category, we also ensure that the total number of true and false images in the cached model is balanced, that is, we extract the same real and fake images from each dataset. It also simulates a simulated real mixed environment, that is, the number of real images is greater than that of a single generative model.Specifically, the few images we extract from these datasets are regarded as the support set $\mathcal{D}_{support} = \{(x_i, y_i)\}_{i=1}^{N}$ for building our cached model.A set of images to be tested is regarded as a query set $\mathcal{D}_{query} = \{(x_q)\}_{q=1}^{N_q}$.This task aims to predict the category of each image in the query set based on the information provided by the support set.

We aim to enhance the model's capability to detect synthetic images by integrating a small amount of new knowledge, eliminating the need for traditional large-scale training. Leveraging a pre-trained CLIP model and a limited number of dataset samples from the target domain, we strategically extract image features from the intermediate layers of the CLIP to effectively capture forgery traces in synthetic images. Our proposed cache model comprises two core components: a Feature Bank and a Label Bank. Specifically, we utilize the CLIP to extract $D$-dimensional features normalized via $\textit{L}_2$ normalization $f_k \in \mathbb{R}^D$. The true label $y_k \in \{\text{real}, \text{fake}\}$ corresponding to each image $x_k$ is converted into a 2-dimensional one-hot encoding vector $l_k \in \mathbb{R}^2$. The $f_k$ and $l_k$ are as follows:

\begin{equation}
    f_k = \frac{E_{vis}(x_k)}{\| E_{vis}(x_k) \|}
\end{equation}

\begin{equation}
    l_k = \text{OneHot}(y_k)
\end{equation}

These extracted intermediate layer features \( f_k \) are treated as keys, aggregated into Feature Bank. 
The one-hot ground-truth vectors \( l_k \) are used as their values, aggregated into Label Bank. 
In the form of key-value pairs, the entire cache model can be expressed as 
$(\mathcal{K}, \mathcal{V}) = ( \{f_k\}_{k=1}^{N_S}, \{l_k\}_{k=1}^{N_S} )$
This cached model is built before the inference phase begins; in this way, the model remembers all new knowledge extracted from a few samples.

\subsection{Few-shot Training-free Network (FTNet) }
For an unseen test image, we use the CLIP intermediate layer to extract global features $f_{test}$ , which are in the same embedding space as the features in the feature bank. We only need two simple matrix-vector multiplications to achieve classification. First, we can calculate the cosine similarity between the test image features and $F_{B}$ by 

\begin{equation}
    S_{\cos} = \frac{f_{\text{test}} F_{B}^T}{\| f_{\text{test}} \| \cdot \| F_{B} \|} \in \mathbb{R}^{1 \times N}
\end{equation}

This represents the semantic relevance of the test image to the images in the cache. 
Then, use \( S_{\cos} \) as the weight to integrate the one-hot encoded labels in the label bank $L_{B}$ to obtain the classification  \( \text{logits} \in \mathbb{R}^{1 \times K} \). The process can be formalized as follows:

\begin{equation}
\mathrm{logits} = \varphi(S_{\cos} {L}_{B})
\end{equation}

\renewcommand{\arraystretch}{1.2}
\setlength{\tabcolsep}{4pt} 
\begin{table*}[!t]
\centering
\resizebox{\textwidth}{!}{ 
\begin{tabular}{c|c|*{6}{c}|c|*{6}{c}|c}
\toprule
\multirow{2}{*}{Method} & \multirow{2}{*}{Reference} 
& \multicolumn{7}{c|}{\textbf{Intra-domain}} 
& \multicolumn{7}{c}{\textbf{Cross-domain}} \\
\cmidrule(lr){3 - 9} \cmidrule(lr){10 - 16} 
& & {Midjourney} & {SD} & {BigGAN} & {ADM} & {VQDM} & {GLIDE} & {\textbf{mAcc}} 
& {Midjourney} & {SD} & {BigGAN} & {ADM} & {VQDM} & {GLIDE} & {\textbf{mAcc}} \\
\midrule
GramNet \cite{liu2020global}   & CVPR2020 & 75.8 & 81.8 & 93.1 & 58.6 & 76.2 & 88.9 & 79.0
                  & 54.2 & 99.0 & 51.7 & 50.3 & 50.8 & 54.6 & 69.9 \\
CNNSpot \cite{wang2020cnn}     & CVPR2020 & 88.0 & 92.3 & 98.9 & 81.2 & 93.1 & 95.8 & 91.5
                    & 52.8 & 90.2 & 46.8 & 50.1 & 53.4 & 39.8 & 64.2 \\
SBI \cite{shiohara2022detecting}       & CVPR2022 & 87.9 & 87.1 & 84.5 & 84.4 & 86.7 & 97.1 & 87.9
                    & 66.8 & 99.2 & 50.0 & 50.6 & 49.8 & 55.8 & 62.0 \\
F3Net \cite{qian2020thinking}       & ECCV2020 & 90.5 & 89.6 & 95.0 & 84.1 & 93.3 & 98.1 & 91.8
                    & 50.1 & 99.9 & 49.9 & 49.9 & 49.9 & 50.0 & 68.7 \\
UnivFD \cite{ojha2023towards}         & CVPR2023 & 85.5 & 85.1 & 89.1 & 96.9 & 86.7 & 88.3 & 88.6
                   & 93.9 & 95.6 & 90.5 & 71.9 & 81.6 & 85.4 & 88.8 \\
NPR \cite{tan2024rethinking}   & CVPR2024 & 89.6 & 91.2 & 98.7 & 92.7 & 91.5 & 95.2 & \textbf{93.1}
                    & 81.0 & 97.6 & 84.2 & 76.9 & 84.1 & 89.8 & 88.6 \\
FreqNet \cite{tan2024frequency}      & AAAI2024 & 85.8 & 77.6 & 98.9 & 90.9 & 84.8 & 89.2 & 87.8
                    & 89.6 & 98.2 & 81.4 & 66.8 & 75.8 & 86.5 & 86.8 \\
AIDE \cite{yan2024sanity}      & ICLR2025 & 90.1 & 90.7 & 95.5 & 92.9 & 87.4 & 93.8 & 91.7
                    & 79.4 & 99.4 & 66.9 & 78.5 & 80.3 & 91.8 & 86.9 \\
                      
\hline
\rowcolor{gray!20} 
FTNet (\textbf{ours})       & 4-shot & 93.8 & 91.8 & 94.5 & 85.8 & 78.7 & 94.6 & 89.9
                   & 93.4 & 86.9 & 97.1 & 90.0 & 80.2 & 96.6 & \underline{90.7} \\
\rowcolor{gray!20} 
FTNet-T (\textbf{ours})       & 4-shot & 94.9 & 92.9 & 95.6 & 90.7 & 84.1 & 95.5 & \underline{92.3}
                   & 92.1 & 93.1 & 96.5 & 94.7 & 92.5 & 96.1 & \textbf{94.2} \\
\bottomrule
\end{tabular}
} 
\caption{Comparison under a Single-Source Training Scheme. In cross-domain testing, results are taken from the original papers \cite{tan2025c2p,yan2024sanity}. The baseline methods are trained on the SDv1.4 dataset. In intra-domain testing, results are from re-implementation. Bold and \underline{underline} indicate the best and second-best performances, respectively.}
\label{tab:intra-cross}
\end{table*}

\renewcommand{\arraystretch}{1.2}
\setlength{\tabcolsep}{1pt}


\begin{table}[t]
\centering
\scalebox{0.75}{
\small 
\begin{tabular}{c|c|cccccc|c}
\toprule
Method & Ref & Midjourney & SD & BigGAN & ADM & VQDM & GLIDE & \textbf{mAcc} \\
\midrule
GramNet (\citeyear{liu2020global})   & CVPR20 & 58.1 & 72.8 & 61.2 & 58.7 & 57.8 & 65.3 & 62.3 \\
CNNSpot (\citeyear{wang2020cnn})   & CVPR20 & 58.2 & 70.3 & 56.6 & 57.0 & 56.7 & 57.1 & 59.3 \\
F3Net (\citeyear{qian2020thinking})     & ECCV20 & 55.1 & 73.1 & 56.5 & 66.5 & 62.1 & 57.8 & 61.9 \\
UniFD (\citeyear{ojha2023towards})         & CVPR23 & 70.8 & 74.6 & 86.1 & 70.0 & 71.9 & 73.2 & 74.4 \\
DIRE (\citeyear{wang2023dire})     & ICCV23 & 65.0 & 73.7 & 56.7 & 61.9 & 63.4 & 69.1 & 65.0 \\
LARE2 (\citeyear{luo2024lare})     & CVPR24 & 66.4 & 87.3 & 74.0 & 66.7 & 84.4 & 81.3 & 76.7 \\
NPR (\citeyear{tan2024rethinking})     & CVPR24 & 74.6 & 76.3 & 83.2 & 70.6 & 64.9 & 89.2 & 76.5 \\
FSD(zero-shot) (\citeyear{wu2025few})     & ICML25 & 75.1 & 88.0 & 62.1 & 74.1 & 69.1 & 93.9 & 77.1 \\
\midrule
FSD(10-shot) (\citeyear{wu2025few})      & ICML25 & 80.9 & 88.8 & 82.2 & 79.2 & 76.2 & 97.1 & 84.1 \\ 
\cmidrule(lr){1-9}
\cellcolor{gray!20}FTNet (\textbf{ours})       & \cellcolor{gray!20}4-shot & \cellcolor{gray!20}93.4 & \cellcolor{gray!20}86.9 & \cellcolor{gray!20}97.1 & \cellcolor{gray!20}90.0 & \cellcolor{gray!20}80.2 & \cellcolor{gray!20}96.6 & \cellcolor{gray!20}\underline{90.7} \\
\cellcolor{gray!20}FTNet-T (\textbf{ours})       & \cellcolor{gray!20}4-shot & \cellcolor{gray!20}92.1 & \cellcolor{gray!20}93.1 & \cellcolor{gray!20}96.5 & \cellcolor{gray!20}94.7 & \cellcolor{gray!20}92.5 & \cellcolor{gray!20}96.1 & \cellcolor{gray!20}\textbf{94.2} \\
\bottomrule
\end{tabular}
}
\caption{Comparison with Cross-Generator Validation Scheme. We implemented UniFD and NPR while other baselines are directly cited from \cite{wu2025few}.}
\label{tab:cross}
\end{table}

where \( \varphi(x) = \exp(-\alpha(1 - x)) \) is an activation function\cite{zhang2021tip}. \( \alpha \) represents a tuning hyperparameter. The exponential function converts the similarity into a non-negative value and uses \( \alpha \) to modulate its sharpness. In \( S_{\cos} \) similar feature memories with higher scores contribute more to the final classification logits, and vice versa. Through this similarity-based label integration, FTNet can adaptively distinguish synthesized images.

\subsection{FTNet with Fine-tuning (FTNet-T)}
FTNet can quickly identify new AI synthetic images by caching a small amount of target domain sample knowledge. However, as the number of cached samples increases, the generalization ability of the model tends to be flat. In order to further improve the generalization ability, we propose the FTNet-T, which fine-tunes a small number of samples in the cache, that is, fine-tunes the keys in the Feature Bank. In view of the extreme scarcity of new generative model samples in the real world, FTNet-T achieves advanced performance on advanced forged image datasets with a fine-tuning cost as low as 20 epochs.

Specifically, we unfreeze the keys in the Feature Bank, keep the values in the Label Bank frozen, and add a linear layer $L_A(\cdot)$ to perform learning. The similarity is calculated by the output of the linear layer. The formula is as follows:

\begin{equation}
   S_{\cos} = L_A(f_{test}) = f_{test} W_A
\end{equation}

Its internal learnable weight matrix is $\mathbf{W}_A \in \mathbb{R}^{D \times N}$.
$\mathbf{W}_A$ is initialized as $F_{\text{B}}^T$. However, it will be updated according to the task objectives. During fine-tuning, cross-entropy loss is used as the main objective function to measure the difference between the model prediction and the true label.

\begin{table*}[t]
\small
\centering
\resizebox{\textwidth}{!}{%
\renewcommand{\arraystretch}{1.2} 
\setlength{\tabcolsep}{4pt} 
\begin{tabular}{
c
|
c
|
*{6}{c}
c
*{2}{c}
*{2}{c}
c
*{3}{c}
*{3}{c}
c
|
c
}
\toprule
\multirow{2}{*}[-4pt]{Method}
& \multirow{2}{*}[-4pt]{Reference} 
& \multicolumn{6}{c}{GAN} 
& \multicolumn{1}{c}{\multirow{2}{*}[-4pt]{\makecell{Deep\\fakes}}}
& \multicolumn{2}{c}{Low level} 
& \multicolumn{2}{c}{Perceptual loss} 
& \multicolumn{1}{c}{\multirow{2}{*}[-4pt]{Guided}}
& \multicolumn{3}{c}{LDM} 
& \multicolumn{3}{c}{Glide} 
& \multicolumn{1}{c}{\multirow{2}{*}[-4pt]{Dalle}}
& \multicolumn{1}{|c}{\multirow{2}{*}[-4pt]{\textbf{mAcc}}}\\
\cmidrule(lr){3-8} \cmidrule(lr){10-11} \cmidrule(lr){12-13} \cmidrule(lr){15-17} \cmidrule(lr){18-20}
&  
& \multicolumn{1}{c}{\makecell{Pro\\GAN}} & \multicolumn{1}{c}{\makecell{Cycle\\GAN}} & \multicolumn{1}{c}{\makecell{Big\\GAN}} & \multicolumn{1}{c}{\makecell{Style\\GAN}} & \multicolumn{1}{c}{\makecell{Gau\\GAN}} & \multicolumn{1}{c}{\makecell{Star\\GAN}} 
& \multicolumn{1}{c}{}
& \multicolumn{1}{c}{SITD} & \multicolumn{1}{c}{SAN}
& \multicolumn{1}{c}{CRN} & \multicolumn{1}{c}{IMLE}
& \multicolumn{1}{c}{}
& \multicolumn{1}{c}{\makecell{200\\steps}} & \multicolumn{1}{c}{\makecell{200\\w/cfg}} & \multicolumn{1}{c}{\makecell{100\\steps}}
 & \multicolumn{1}{c}{\makecell{100\\27}} & \multicolumn{1}{c}{\makecell{50\\27}} & \multicolumn{1}{c}{\makecell{100\\10}} && \\
\midrule
CNN-Spot(\citeyear{wang2020cnn}) & CVPR2020 
& 99.99 & 85.20 & 70.20 & 85.70 & 78.95 & 91.70 & 53.47 & 66.67 & 48.69 & 86.31 & 86.26 & 60.07 & 54.03 & 54.96 & 54.14 & 60.78 & 63.80 & 65.66 & 55.58 & 69.58 \\
Patchfor (\citeyear{chai2020makes}) &ECCV2020  
& 75.03 & 68.97 & 68.47 & 79.16 & 64.23 & 63.94 & 75.54 & 75.14 & 75.28 & 72.33 & 55.30 & 67.41 & 76.50 & 76.10 & 75.77 & 74.81 & 73.28 & 68.52 & 67.91 & 71.24 \\
Freq-spec (\citeyear{zhang2019detecting}) &WIFS2019  
& 49.90 & 99.90 & 53.50 & 49.90 & 50.30 & 99.70 & 50.10 & 50.00 & 48.00 & 50.60 & 50.10 & 50.90 & 50.40 & 50.40 & 50.30 & 51.70 & 51.40 & 50.40 & 50.00 & 55.45
\\
F3Net (\citeyear{qian2020thinking}) &ECCV2020  
& 99.38 & 76.38 & 65.33 & 92.56 & 58.10 & 100.00 & 63.48 & 54.17 & 47.26 & 51.47 & 51.47 & 69.20 & 68.15 & 75.35 & 68.80 & 81.65 & 83.25 & 83.05 & 66.30 & 71.33 \\
UniFD (\citeyear{ojha2023towards}) &CVPR2023  
& 100.00 & 98.50 & 94.50 & 82.00 & 99.50 & 97.00 & 66.60 & 63.05 & 57.50 & 59.50 & 72.00 & 70.03 & 94.19 & 73.76 & 94.36 & 79.07 & 79.85 & 78.14 & 78.68 & 81.38 \\
LGrad (\citeyear{tan2023learning}) &CVPR2023  
& 99.84 & 85.39 & 82.88 & 94.83 & 72.45 & 99.62 & 58.00 & 62.50 & 50.00 & 50.74 & 50.78 & 77.50 & 94.20 & 95.85 & 94.80 & 87.40 & 90.70 & 89.55 & 88.35 & 80.28 \\
FreqNet (\citeyear{tan2024frequency}) &AAAI2024  
& 97.90 & 95.84 & 90.45 & 97.55 & 90.24 & 93.41 & 97.40 & 88.82 & 59.04 & 71.92 & 67.35 & 86.70 & 84.55 & 99.58 & 65.56 & 85.69 & 97.40 & 88.15 & 59.06 & 85.09 \\
NPR (\citeyear{tan2024rethinking}) &CVPR2024  
& 99.84 & 95.00 & 87.55 & 96.23 & 86.57 & 99.75 & 76.90 & 66.94 & 98.63 & 50.00 & 50.00 & 84.55 & 97.65 & 98.00 & 98.20 & 96.25 & 97.15 & 97.35 & 87.15 & 87.56 \\
FatFormer (\citeyear{liu2024forgery}) & CVPR2024 
& 99.89 & 99.32 & 99.50 & 97.15 & 99.41 & 99.75 & 93.23 & 81.11 & 68.04 & 90.45 & 90.45 & 76.00 & 98.60 & 94.90 & 98.65 & 94.35 & 94.65 & 94.20 & 98.75 & 90.86 \\
C2p-clip (\citeyear{tan2025c2p}) &AAAI2025  
& 99.71 & 90.69 & 95.28 & 99.38 & 95.26 & 96.60 & 89.86 & 98.33 & 64.61 & 90.69 & 90.69 & 77.80 & 99.05 & 98.05 & 98.05 & 94.65 & 94.20 & 94.40 & 98.80 & 93.00 \\
\hline
\rowcolor{gray!20} 
FTNet (\textbf{ours}) & 4-shot 
& 98.37 & 96.17 & 76.18 & 99.55 & 75.63 & 100.00 & 83.38 & 83.24 & 75.58 & 99.97 & 99.97 & 98.80 & 100.00 & 100.00 & 100.00 & 99.85 & 99.90 & 100.00 & 100.00 & \underline{94.03} \\
\rowcolor{gray!20} 
\rowcolor{gray!20} 
FTNet-T (\textbf{ours}) & 4-shot 
& 99.69& 99.20 & 97.47 & 99.77 & 97.85 & 100.00 & 85.45 & 98.58 & 76.05 & 99.95 & 99.95 & 93.17 & 100.00 & 99.95 & 100.00 & 99.35 & 99.55 & 99.60 & 99.95 & \textbf{97.14} \\
\bottomrule
\end{tabular}%
} 
\caption{Comparison on the UniversalFakeDetect. The baseline results are directly cited from \cite{tan2025c2p}. All baseline methods are trained on the ProGAN dataset. Bold and \underline{underline} indicate the best and second-best performances, respectively.}
\label{tab:cross3}
\end{table*}

\section{Experiments}

In this section, we evaluate the performance of our FTNet and FTNet-T methods for few-shot deepfake detection through a series of experiments. We first describe the datasets and implementation settings, then compare their performance with baseline detectors across various datasets, and finally conduct ablation studies to assess the impact of critical factors such as sample count and CLIP layers.

\renewcommand{\arraystretch}{1.1}
\setlength{\tabcolsep}{7pt}
\begin{table*}[ht] 
\centering
\resizebox{\textwidth}{!}{%
\small 
\begin{tabular}{c|c|cc|cc|cc|cc|cc|cc} 
\toprule
\multirow{2}{*}[-3pt]{Method} & \multirow{2}{*}[-3pt]{Reference} & \multicolumn{2}{c}{SD1.5} & \multicolumn{2}{|c}{SD2.1} & \multicolumn{2}{|c}{SDXL} & \multicolumn{2}{|c}{SD3} & \multicolumn{2}{|c}{Flux.1} & \multicolumn{2}{|c}{\textbf{AVG}} \\
 \cmidrule(lr){3-4}\cmidrule(lr){5-6}\cmidrule(lr){7-8}\cmidrule(lr){9-10}\cmidrule(lr){11-12}\cmidrule(lr){13-14}
 &  &  F1&Acc &F1&Acc &F1 & Acc&F1 & Acc&F1 & Acc&F1 & mAcc \\
\midrule
GramNet \cite{liu2020global} & CVPR2020 & 80.51 & 80.35 & 74.01 & 76.66 & 65.28 & 70.76 & 64.35 & 70.29 & 52.00 & 63.37 & 67.23 & 72.29 \\
CNNSpot \cite{wang2020cnn} & CVPR2020 & 84.60 & 85.04 & 71.56 & 75.94 & 59.70 & 68.72 & 56.27 & 67.08 & 35.72 & 57.57 & 61.57 & 70.87 \\
MVSS-Net \cite{chen2021image} & CVPR2021  & 93.47 & 93.65 & 79.27 & 82.33 & 59.85 & 70.42 & 62.80 & 72.13 & 27.59 & 56.78 & 64.60 & 75.06 \\
ObjectFormer \cite{wang2022objectformer} & CVPR2022 & 71.72 & 75.22 & 66.79 & 72.55 & 49.19 & 62.92 & 48.32 & 62.54 & 37.92 & 58.05 & 54.79 & 66.26 \\

UniFD \cite{ojha2023towards} & CVPR2023 & 77.45 & 77.60 & 80.62 & 81.92 & 70.74 & 74.83 & 71.09 & 75.17 & 61.10 & 69.06 & 72.20 & 75.72 \\
FreqNet \cite{tan2024frequency} & AAAI2024 & 75.88 & 77.70 & 60.97 & 68.37 & 53.15 & 64.02 & 53.50 & 64.37 & 38.47 & 57.08 & 56.39 & 66.31 \\
IML-ViT \cite{ma2023iml} & AAAI2024 & 94.47 & 75.73 & 69.70 & 61.19 & 40.98 & 49.95 & 44.69 & 51.25 & 18.20 & 43.62 & 53.61 & 56.35 \\
NPR \cite{tan2024rethinking} & CVPR2024 & 79.41 & 79.28 & 81.67 & 81.84 & 72.12 & 74.28 & 73.43 & 75.47 & 67.62 & 71.36 & 74.85 & 76.45 \\
RINE \cite{koutlis2024leveraging} & ECCV2024 & 91.08 & 90.98 & 87.47 & 88.12 & 73.43 & 78.76 & 72.05 & 76.78 & 55.86 & 67.02 & 75.98 & 80.33 \\
MaskCLIP \cite{wang2025opensdi} & CVPR2025& 92.64 & 92.72 & 88.71 & 89.45 & 78.02 & 81.22 & 73.07 & 78.01 & 56.49 & 68.50 & 77.79 &\underline{81.98} \\
\hline
\rowcolor{gray!20} 
FTNet (\textbf{ours}) & 8-shot & 73.24 & 74.62 & 80.17 & 82.21 & 82.14 & 83.63 & 76.58 & 79.64 & 82.57 & 79.58 &\underline{77.83} & 79.94 \\
\rowcolor{gray!20} 
FTNet-T (\textbf{ours}) & 8-shot & 78.04 & 77.70 & 84.13 & 84.79 & 85.82 & 86.17 & 84.06 & 84.69 & 81.39 & 82.54 & \textbf{82.68} & \textbf{83.16} \\
\bottomrule
\end{tabular}
}
\caption{Comparison on the OpenSDI. The baseline methods results are directly cited from paper\cite{wang2025opensdi}. The baseline methods compared are all trained on the SD1.5 dataset (100k real images and 100k fake images).}
\label{tab:cross1}
\end{table*}

\subsection{Settings}


\noindent\textbf{Datasets:} To ensure the consistency of the benchmark, we used 3 benchmark datasets for implementation. They are \textbf{GenImage}, \textbf{UniversalFakeDetect}, and \textbf{OpenSDI} dataset.
The GenImage million-level dataset contains 8 generators (7 Diffusion models and one GAN).The generators include Midjourney\cite{midjourney2022}, Stable Diffusion 
V1.4\cite{Rombach_2022_CVPR}, StableDiffusionV1.5\cite{Rombach_2022_CVPR}, Wukong\cite{wukong2022}, ADM\cite{dhariwal2021diffusion}, GLIDE\cite{nichol2021glide}, VQDM\cite{gu2022vector} and BigGAN\cite{brock2018large}. 
The UniversalFakeDetect\cite{ojha2023towards} includes 20 subsets of generated images. The test set comprises 19 subsets from various generative models including ProGAN\cite{karras2017progressive}, StyleGAN\cite{karras2019style}, BigGAN\cite{brock2018large}, CycleGAN\cite{zhu2017unpaired}, StarGAN\cite{choi2018stargan}, GauGAN\cite{park2019semantic}and Deepfake\cite{rossler2019faceforensics++}, CRN\cite{chen2017photographic}, IMLE\cite{li2019diverse}, SAN\cite{dai2019second}, SITD\cite{chen2018learning}, Guided\cite{dhariwal2021diffusion}, LDM\cite{Rombach_2022_CVPR},
Glide\cite{nichol2021glide}, DALLE\cite{ramesh2021zero}. The OpenSDI\cite{wang2025opensdi} open world complex dataset contains 5 generative models. The generators include SD1.5\cite{Rombach_2022_CVPR}, SD2.1\cite{Rombach_2022_CVPR}, SDXL\cite{podell2023sdxl}, SD3\cite{esser2024scaling} and Flux.1\cite{flux2024}.
Since we regard different generative models as different categories and real images as a separate category, following the settings of Wu et al. \cite{wu2025few}, the three diffusion models SDv1.4, SDv1.5 and WuKong in the GenImage have the same model structure and are difficult to distinguish, so the three models are unified into the SD model. To replicate the data of previous work, we average the acc indicators of the three models to obtain the acc of SD.

\noindent\textbf{Implementation Details:} Our experiments are conducted under 4-shot and 8-shot few-shot learning configurations. We define k-shot as follows: to build the cache model, we randomly sample k real and k fake images from each generative model's dataset. This sampling strategy ensures that the total number of real images in the cache (summed across all sources) is greater than the number of fake images from any single generator, which closely simulates real-world scenarios. For our model architecture, we utilize a pretrained CLIP model with a ViT-L/14 backbone and set the input image resolution to 224x224 pixels. Regarding hyperparameters, the temperature parameter  \( \alpha \) in the FTNet module's activation function is set to 15. For the fine-tunable FTNet, we employ an AdamW \cite{kingma2014adam} optimizer with a learning rate of 0.001. The primary evaluation metric is Accuracy (Acc). Additionally, to ensure a comprehensive comparison on the OpenSDI dataset \cite{wang2025opensdi}, we also report the F1-score. All of our methods are implemented using PyTorch and run on an Nvidia GeForce RTX 3080 GPU.

\subsection{Comparisons with State-of-the-art Methods}
\noindent\textbf{Evaluation on Genimage.} 
To fully evaluate our framework on the GenImage dataset, we adopt two common schemes, as shown in Table \ref{tab:intra-cross} and Table \ref{tab:cross}. These schemes are designed to evaluate different aspects of the model's generalization ability, from standard single-source migration to strict cross-generator robustness.
The results of accuracy (Acc) are presented. 
First, Table \ref{tab:intra-cross} evaluates the single-source generalization ability, including two benchmarks: in-domain and cross-domain testing. In the in-domain experiment, the training and test generator types are consistent. We reduced the training set of the baseline method. Each method's training set is 2K images (real and fake ratio 1:1), and the official experimental setting is used for training. In our intra-domain experiment, we randomly sample from the original training set. Our FTNet-T achieves excellent results in the 4-shot intra-domain test. Its average accuracy reaches 92.3\%, ranking second among all methods and significantly improving the performance by 3.7\% compared to the baseline method UniFD. In the cross-domain test, all baseline methods are trained on the original SDv1.4 dataset, while our method bypasses large-scale training and only needs a small number of samples from the target domain to achieve SOTA performance. In the 4-shot setting, our method FTNet improves 1.9\% over UniFD, and 5.4\% over the baseline when we further fine-tune. These results highlight that our method can be directly applied to detection tasks without any training or parameter updates, and its performance is superior to traditional methods that rely on large-scale data training.
Table \ref{tab:cross} adopts the cross-generator verification scheme. This scheme trains 6 specialized models on 6 independent generators for the baseline method and calculates their average performance on unseen data. This scheme clearly reveals the limitations of traditional zero-shot methods when facing unknown generative models; that is, when the test data differs too much from the training source, its performance will seriously decline. As shown in the results in the table, our method effectively alleviates this problem: untrained FTNet improves by 6.6\% compared to FSD (10-shot), while FTNet-T improves by 10.1\%. This shows that our method can effectively alleviate the negative impact caused by the difference between training and test domains.

\begin{figure}[!t]

  \centering
   \includegraphics[width=\columnwidth]{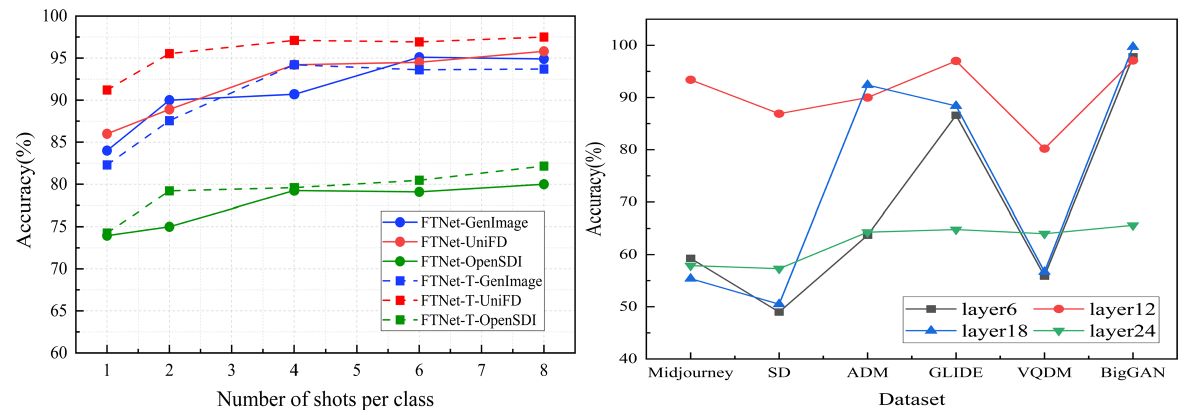}
   \vspace{-0.1in}
   \caption{Analysis of the method's performance. (Left) Influences of the number of shots. (Right) Performance of the FTNet method on different CLIP layers.}
   \vspace{-0.1in}
   \label{fig:fig-few}
   
\end{figure}


\begin{table}[htbp]
\centering
\small  
\setlength{\tabcolsep}{4pt}  
\renewcommand{\arraystretch}{1.1}  

\resizebox{\linewidth}{!}{
\begin{tabular}{l|c| c c c|c}  
\toprule
Method & Sampling & GenImage & UniFD & OpenSDI & \textbf{mAcc} \\
\midrule
UniFD (\citeyear{ojha2023towards})         & 8-shot & 70.0  & 85.9 & 52.1 & 69.3 \\
NPR (\citeyear{tan2024rethinking})           & 8-shot & 78.8  & 87.4 & 55.1 & 73.8 \\
\hline
FTNet (\textbf{ours})      & 4-shot & 90.7  & 89.3 & 79.3 & 86.4 \\
FTNet-T (\textbf{ours})      & 4-shot & \textbf{94.2}  & \textbf{97.1} & \textbf{79.6} & \textbf{90.3} \\
\bottomrule
\end{tabular}
}
\caption{Comparison with fine-tuning-based methods.} 
\vspace{-0.1in}

\label{tab:experiments}   

\end{table}

\noindent\textbf{Evaluation on UniversalFakeDetect.}
The accuracy (Acc) results are shown in Table\ref{tab:cross3}. UniFD is similar to our method in that it retains the original pre-trained knowledge of CLIP and only achieves classification by fine-tuning the fully connected layer. Compared with UniFD, our method without training improves mAcc by 12.65\%. When we adopt light fine-tuning, mAcc improves by 15.76\%. In addition, compared to the latest state-of-the-art method, C2P-clip, our method improves accuracy by 4.14\%, which proves the superiority of our method without the need for text encoders and large sample training.

\noindent\textbf{Evaluation on OpenSDI.} 
The accuracy (Acc) results are listed in Table \ref{tab:cross1}. The OpenSDI is designed to simulate the image forgery detection challenge in the open world. The construction of this dataset fully considers the three core open-world settings of ``user diversity'', ``model innovation'', and ``operation range'', making it a more challenging and realistic benchmark. Compared with the baseline UniFD, FTNet improves the accuracy by 4.22\%, and the accuracy is improved by 7.44\% after further fine-tuning. In addition, compared to the MaskClip, FTNet-T improved accuracy by 1.18\%, but MaskClip combines the visual and text encoders of CLIP and a masked autoencoder (MAE)\cite{he2022masked} encoder, which relies on high-performance computing facilities and is very time-consuming.

\subsection{Ablation Study}
\noindent\textbf{The number of shots.}
To determine the minimum number of samples required for this method to achieve the best balance between performance and computational overhead, we conduct an ablation experiment. We examine the corresponding performance of the model when taking different sampling numbers. We test it on three major datasets, named GenImage, UniFD (UniversalFakeDetect), and OpenSDI datasets. The experimental results are shown in Figure \ref{fig:fig-few} (left). The performance improves with the increase in the number of samples, but this growth is not linear. When the number of samples reaches a certain scale, the performance gain becomes very limited. Taking all factors into consideration, our method has the best overall performance under 4-shot sampling.

\noindent\textbf{Few-shot Evaluation with Fine-tuning Baselines.}
To rigorously evaluate the performance of our approach, we conducted a comparative experiment. We fine-tuned the baseline models UniFD and NPR using 8-shot training and compared the results with our approach, which utilized only 4-shot samples, as shown in Table \ref{tab:experiments}. The experimental results show that, even with only 4-shot training samples, our FTNet achieves 86.4\% mAcc, significantly outperforming both 8-shot baselines. The fine-tuned FTNet-T further improves performance to 90.3\% mAcc. This result significantly outperforms the 8-shot fine-tuned UniFD and NPR by 21.0\% and 16.5\%, respectively. This comparison fully demonstrates the superior performance of our approach.

\begin{table}[!t]
\small 
\centering
\setlength{\tabcolsep}{2pt}  
\renewcommand{\arraystretch}{1.2}  
\resizebox{\linewidth}{!}{
  \begin{tabular}{l|c|ccccc|c}
  \toprule
  \multirow{2}{*}[-3pt]{Method} & \multirow{2}{*}[-3pt]{\makecell{Training\\samples}} & \multicolumn{5}{c|}{\textbf{Target domain}} & \multirow{2}{*}[-3pt]{\textbf{mAcc}} \\
  \cmidrule(lr){3-7}  
  &  & ADM & BigGAN & GLIDE & VQDM & Midjourney  \\ 
  \midrule
  UniFD (\citeyear{ojha2023towards})    & 320k & 71.9 & \textbf{90.5} & 85.4 & 81.6 & \textbf{93.9} & 84.6 \\
  NPR (\citeyear{tan2024rethinking})      & 320k & 76.9 & 84.2 & 89.8 & \textbf{84.1} & 81.0 & 83.2 \\
  \midrule
  FTNet (\textbf{ours})& 500 & 73.9 & 85.5 & 85.2 & 70.9 & 84.3 & 80.8 \\
  FTNet-T (\textbf{ours})& 500 & \textbf{78.3} & 88.8 & \textbf{94.4} & 75.8 & 92.5 & \textbf{86.0} \\ %
  \bottomrule
  \end{tabular}
}
\caption{Comparison with zero-shot generalization methods.}
\label{table:zero-shot}
\end{table}

\noindent\textbf{Zero-shot Generalization Performance Comparison.}
To evaluate the zero-shot generalization capability of our method, we compare it with two baseline methods, UniFD and NPR, and the results are shown in Table \ref{table:zero-shot}. Our model shows a significant advantage in data efficiency. Specifically, both UniFD and NPR are trained on the full SDv1.4 dataset (320K images in total), while our FTNet-T is adaptively fine-tuned on only 0.15\% of the data (500 images), achieving an improvement in accuracy (mAcc) of about 1.4\% and 2.8\% over UniFD and NPR, respectively. This highlights the strong potential of our method to learn universal forgery patterns in resource-constrained scenarios.

\begin{figure}[!t]
\centering
\includegraphics[width=0.8\columnwidth]{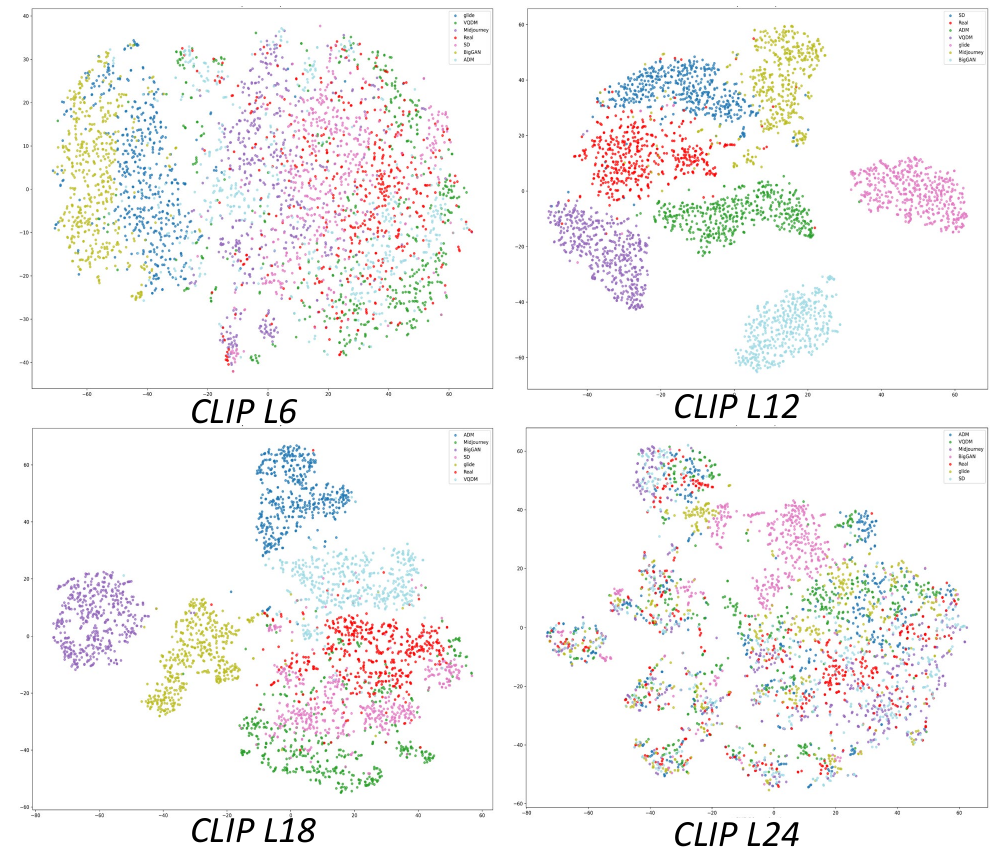}
\caption{Visualization of FTNet Using CLIP ViT-L/14 Features (L6, L12, L18, L24) on the GenImage Dataset.}
\label{fig-tsne}
\end{figure}

\noindent\textbf{Generalization Ability of CLIP Feature Layers.}

We explore how different feature layers within the CLIP affect the generalization ability of our method. We utilize FTNet and analyze features from four different layers: the 6th, 12th, 18th, and 24th layers (L6, L12, L18, L24) in a 4-shot setting. The results are shown in Figure \ref{fig:fig-few} (right). The detection performance is not simply linear with the layer depth. Among them, the middle-level features from L12 achieve the highest average accuracy , outperforming all other layers. The task of forged image detection is highly dependent on the perception of unique textures, frequencies, and artifacts left by generative models. CLIP’s L12 strikes the best balance between preserving these critical low-level artifact information and learning high-level representations that distinguish between different generator models.

To visually verify these findings, the t-SNE of the features is shown in Figure \ref{fig-tsne}, where red represents real images and other colors represent different generative models. The feature clusters at L12 are the clearest and most compact, while the features at the shallower L6 are mixed. The excessively deep L24 is overly abstract, making it nearly impossible to distinguish between genuine and forged features. This finding also suggests that excessively deep feature extraction can actually compromise the separability of features required for specific tasks, such as deepfake detection.

\begin{figure}[!t]
  \centering
   \includegraphics[width=\columnwidth]{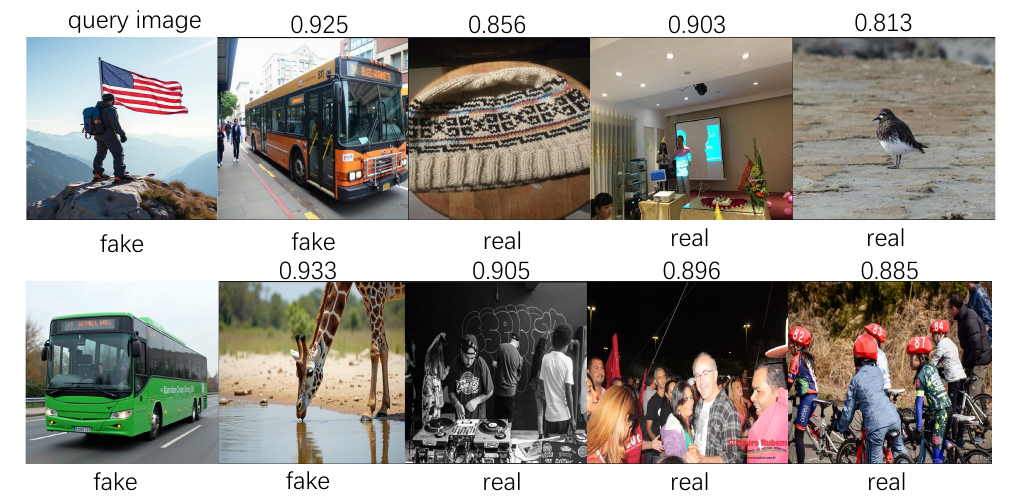}
   \caption{The top and bottom rows display two query cases and their nearest neighbors in the feature space, respectively. The fake query images (e.g., astronaut, bus) are closest to other fake images (e.g., bus, giraffe), despite belonging to completely different semantic categories.}
   \vspace{-0.1in}
   \label{fig:fig-7}
\end{figure}



\noindent\textbf{Feature Space Visualization and Nearest Neighbor Analysis.}
To demonstrate the decision-making process of our method, we conduct a qualitative analysis. As shown in the Figure \ref{fig:fig-7}. We use a dataset from the model Flux.1 and randomly selected a small number of images to construct a support set. The experiments revealed an interesting phenomenon: when a forged query image (e.g., an astronaut) is input, its nearest neighbor in feature space is a completely unrelated forged image (e.g., a bus or a giraffe), rather than a real image with similar content. This finding strongly demonstrates that the CLIP intermediate-level features relied upon by our method successfully capture a universal forgery signature across semantic content, rather than just superficial semantic similarity. This further demonstrates the enormous potential of large-scale pre-trained models to exploit universal features without fine-tuning.


\section{Conclusion}

In this work, we address a critical challenge in deepfake detection by rethinking the task as a few-shot problem rather than a traditional ``zero-shot" task. While many studies treat unseen sample detection as a generalization issue, we highlight that real-world scenarios demand more effective utilization of limited samples, improving performance on previously unseen deepfakes. Our proposed Few-shot Training-free Network (FTNet) provides \textbf{a novel solution by leveraging only one fake sample from the evaluation set}, mimicking real-world conditions where new samples are gathered without requiring retraining or parameter updates. FTNet outperforms traditional methods that depend on large-scale training datasets and achieves state-of-the-art performance, with an 8.7\% improvement on average compared to existing methods. By incorporating this new perspective on few-shot detection, we show that effectively using failed samples in real-world deepfake detection can significantly improve performance. This work provides a step toward more practical solutions for detection in dynamic environments.






\newpage
\clearpage  
\section{Supplementary Materials}

\subsection{Related Work} AI-generated image detection is an important research direction in current digital media, aiming to distinguish between real and highly realistic images synthesized by advanced generative models such as GANs and diffusion models. At present, many works focus on the intrinsic features of images to identify traces left by generative models, which can be roughly divided into three categories: image-based detection methods \cite{marra2019gans, korshunov2022custom}, methods based on frequency domain feature analysis \cite{frank2020leveraging, durall2020watch}, and methods based on spatial domain and semantic features \cite{ojha2023towards, liu2024forgery, tan2025c2p}. For example, to improve generalization to unseen data, Wang et al. \cite{wang2020cnn} employ diverse data augmentation techniques and incorporate large-scale GAN-generated images into the training set. NPR \cite{tan2024rethinking} proposes a simple yet effective artifact representation based on the insight that common up-sampling operations produce generalizable, spatial-domain forgery artifacts. Some works focus on the frequency domain characteristics of images. For example, F3Net \cite{qian2020thinking} enhances the performance of the model by analyzing the division of frequency components and the difference in frequency statistics between real images and forged images. FreqNet \cite{tan2024frequency} aims to overcome the problem of insufficient generalization ability of traditional frequency methods when facing diverse generative models by performing frequency domain learning inside CNN classifiers and focusing on high-frequency information. 
AEROBLADE, proposed by Ricker et al.\cite{ricker2024aeroblade}, is a training-free detection method that leverages the autoencoder (AE) inherent in the latent diffusion model (LDM). Its principle is based on a core observation: the AE reconstruction error of generated images is significantly lower than that of real images. This method achieves detection without training a new classifier by calculating the minimum reconstruction distance of an image among a set of known AEs and making a threshold-based judgment.
With the rise of pre-trained large models, especially visual-language models (VLMs) \cite{radford2021learning}, some studies have begun to take advantage of their powerful feature extraction and semantic understanding capabilities. Ojha et al. \cite{ojha2023towards} added a linear layer to CLIP \cite{radford2021learning} and fine-tuned it with CLIP's powerful feature extraction capabilities. C2P-CLIP \cite{tan2025c2p} efficiently improved the generalization detection performance of the CLIP image encoder by injecting 'category-general prompts' and combining contrastive learning, demonstrating the potential of semantic features in generalization detection. FatFormer \cite{liu2024forgery} enhances detection performance by integrating frequency analysis with a text encoder, which acts as an adapter to the frozen CLIP vision model. Yang et al.\cite{yang2025d} leveraged the CLIP model as a powerful feature extractor and, by introducing an innovative parallel network branch, effectively deconstructed universal fake artifacts present in multi-generator images, thereby significantly enhancing the model’s generalization capability.

The growing diversity and quality of generative models, coupled with the inherent difficulty and high cost of collecting sufficient training data from closed-source models \cite{midjourney2022}, highlight the urgent need for detection methods that can work effectively even with only a small number of new forgery samples. This challenge has greatly stimulated the interest in the few-shot detection paradigm. Wu et al. \cite{wu2025few} are the first to reconceptualize Al-generated image detection as a few-shot classification task, aiming to make detection closer to practical application scenarios. Their proposed few-shot detector (FSD) adopts a novel approach by using a prototypical network \cite{snell2017prototypical} to learn a specialized metric space that can effectively distinguish unseen forged images with very few samples.

\subsection{Dataset Supplement}
Chameleon\cite{yan2024sanity} is a challenging benchmark designed to perform a sanity check on AI-generated image detection. This dataset contains approximately 26,000 high-resolution (720P to 4K) test images. Its key characteristic is that all AI-generated images pass the "Turing Test of Human Perception" and are highly deceptive to human observers. Unlike traditional benchmarks, Chameleon is not limited to a limited number of generators. Instead, it draws on a broad collection of over 11,000 AI-generated images from mainstream AI creation communities such as ArtStation, Civitai, and Liblib, making it more representative of real-world AI image creation. We also used the GANGen dataset\cite{chuangchuangtan-GANGen-Detection}, which is generated by nine different GAN models. There are 4K test images for each model, with equal numbers of real and fake images.

\begin{figure}[!t]
\centering
\includegraphics[width=\columnwidth]{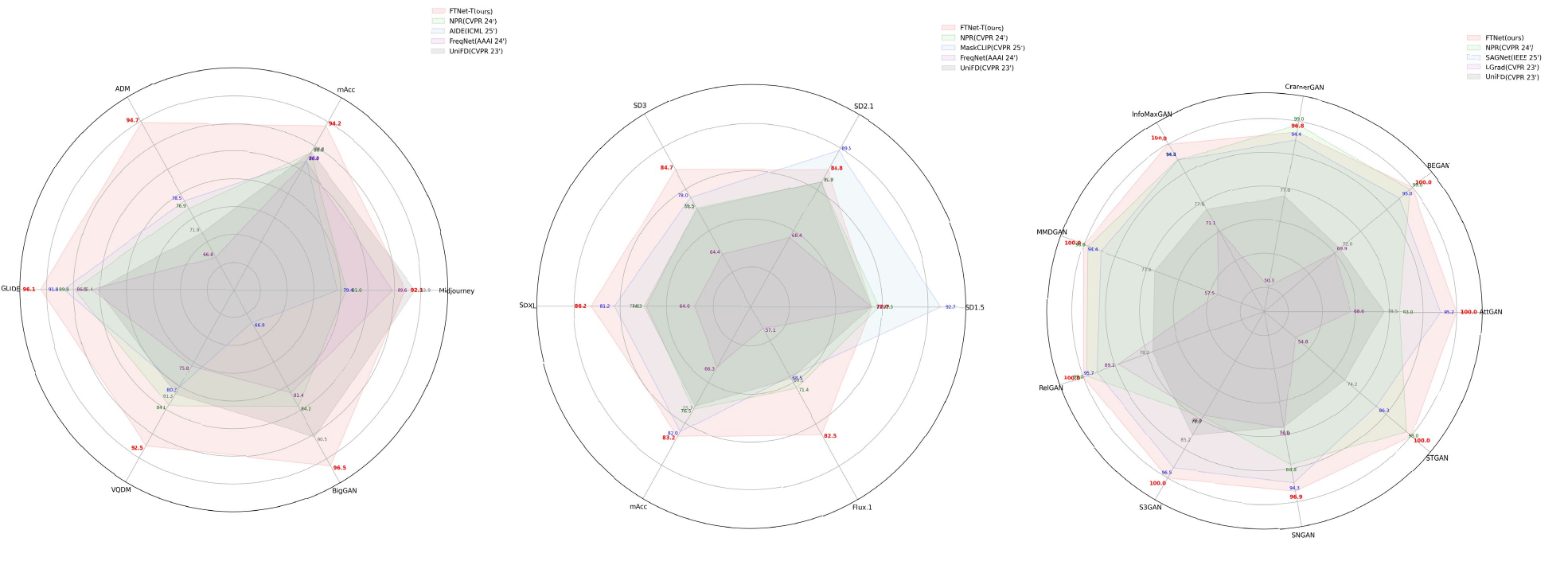}
\caption{Performance comparison with various advanced methods. }
\label{fig-leida}
\end{figure}

\subsection{Comparisons with State-of-the-art Methods}

\noindent\textbf{Evaluation on UniversalFakeDetect} 
 In our manuscript, we present the benchmarking outcomes(acc) of the UniversalFakeDetect Dataset.  The UniversalFakeDetect dataset consists of 19 generators, including GANs, Deepfake, low-level
 vision models, perceptual loss models, and diffusion models. Additionally, we report AP of the dataset. The results are shown in Table \ref{tab:cross4}. Compared to the UniFD, we achieve 98.81\% and 99.20\% mAP , improving 8.67\%. This demonstrates the high practical value of the FTNet framework in the real world, as it can quickly adapt and detect emerging forgery models.

\begin{table*}[t]
\small
\centering
\resizebox{\textwidth}{!}{%
\renewcommand{\arraystretch}{1.2} 
\setlength{\tabcolsep}{4pt} 
\begin{tabular}{
c
|
c
|
*{6}{c}
c
*{2}{c}
*{2}{c}
c
*{3}{c}
*{3}{c}
c
|
c
}
\toprule
\multirow{2}{*}[-4pt]{Method}
& \multirow{2}{*}[-4pt]{Reference} 
& \multicolumn{6}{c}{GAN} 
& \multicolumn{1}{c}{\multirow{2}{*}[-4pt]{\makecell{Deep\\fakes}}}
& \multicolumn{2}{c}{Low level} 
& \multicolumn{2}{c}{Perceptual loss} 
& \multicolumn{1}{c}{\multirow{2}{*}[-4pt]{Guided}}
& \multicolumn{3}{c}{LDM} 
& \multicolumn{3}{c}{Glide} 
& \multicolumn{1}{c}{\multirow{2}{*}[-4pt]{Dalle}}
& \multicolumn{1}{|c}{\multirow{2}{*}[-4pt]{\textbf{mAP}}}\\
\cmidrule(lr){3-8} \cmidrule(lr){10-11} \cmidrule(lr){12-13} \cmidrule(lr){15-17} \cmidrule(lr){18-20}
&  
& \multicolumn{1}{c}{\makecell{Pro\\GAN}} & \multicolumn{1}{c}{\makecell{Cycle\\GAN}} & \multicolumn{1}{c}{\makecell{Big\\GAN}} & \multicolumn{1}{c}{\makecell{Style\\GAN}} & \multicolumn{1}{c}{\makecell{Gau\\GAN}} & \multicolumn{1}{c}{\makecell{Star\\GAN}} 
& \multicolumn{1}{c}{}
& \multicolumn{1}{c}{SITD} & \multicolumn{1}{c}{SAN}
& \multicolumn{1}{c}{CRN} & \multicolumn{1}{c}{IMLE}
& \multicolumn{1}{c}{}
& \multicolumn{1}{c}{\makecell{200\\steps}} & \multicolumn{1}{c}{\makecell{200\\w/cfg}} & \multicolumn{1}{c}{\makecell{100\\steps}}
 & \multicolumn{1}{c}{\makecell{100\\27}} & \multicolumn{1}{c}{\makecell{50\\27}} & \multicolumn{1}{c}{\makecell{100\\10}} && \\
\midrule
CNN-Spot(\citeyear{wang2020cnn}) & CVPR2020 
& 100.0 & 93.47 & 84.5 & 99.54 & 89.49 & 98.15 & 89.02 & 73.75 & 59.47 & 98.24 & 98.4 & 73.72 & 70.62 & 71.0 & 70.54 & 89.65 & 84.91 & 82.07 & 70.59 & 83.58 \\
Patchfor (\citeyear{chai2020makes}) &ECCV2020  
& 80.88 & 72.84 & 71.66 & 85.75 & 65.99 & 69.25 & 76.55 & 76.19 & 76.34 & 74.52 & 68.52 & 75.03 & 87.1 & 86.72 & 86.4 & 85.57 & 83.73 & 78.38 & 75.67 & 77.73 \\
Freq-spec (\citeyear{zhang2019detecting}) &WIFS2019  
& 55.39 & 100.0 & 75.08 & 55.11 & 66.08 & 100.0 & 45.18 & 47.46 & 57.12 & 53.61 & 50.98 & 57.72 & 77.72 & 77.25 & 76.47 & 68.58 & 64.58 & 61.92 & 67.77 & 66.21
\\
F3Net (\citeyear{qian2020thinking}) &ECCV2020  
& 99.96 & 84.32 & 69.90 & 99.72 & 56.71 & 100.0 & 78.82 & 52.89 & 46.70 & 63.39 & 64.57 & 70.53 & 73.76 & 81.66 & 74.62 & 89.81 & 91.04 & 90.86 & 71.84 & 76.89 \\
UniFD (\citeyear{ojha2023towards}) &CVPR2023  
& 100.0 & 98.13 & 94.46 & 86.66 & 99.25 & 99.53 & 91.67 & 78.54 & 67.54 & 83.12 & 91.06 & 79.24 & 95.81 & 79.77 & 95.93 & 93.93 & 95.12 & 94.59 & 88.45 & 90.14 \\
LGrad (\citeyear{tan2023learning}) &CVPR2023  
& 100.0 & 93.98 & 90.69 & 99.86 & 79.36 & 99.98 & 67.91 & 59.42 & 51.42 & 63.52 & 69.61 & 87.06 & 99.03 & 99.16 & 99.18 & 93.23 & 95.10 & 94.93 & 97.23 & 86.35 \\
FreqNet (\citeyear{tan2024frequency}) &AAAI2024  
& 99.92 & 99.63 & 96.05 & 99.89 & 99.71 & 98.63 & 99.92 & 94.42 & 74.59 & 80.10 & 75.70 & 96.27 & 96.06 & 100.0 & 62.34 & 99.80 & 99.78 & 96.39 & 77.78 & 91.95 \\
NPR (\citeyear{tan2024rethinking}) &CVPR2024  
& 100.0 & 99.53 & 94.53 & 99.94 & 88.82 & 100.0 & 84.41 & 97.95 & 99.99 & 50.16 & 50.16 & 98.26 & 99.92 & 99.91 & 99.92 & 99.87 & 99.89 & 99.92 & 99.26 & 92.76 \\
FatFormer (\citeyear{liu2024forgery}) & CVPR2024 
& 100.0 & 100.0 & 99.98 & 99.75 & 100.0 & 100.0 & 97.99 & 97.94 & 81.21 & 99.84 & 99.93 & 91.99 & 99.81 & 99.09 & 99.87 & 99.13 & 99.41 & 99.20 & 99.82 & 98.16 \\
C2p-clip (\citeyear{tan2025c2p}) &AAAI2025  
& 99.99 & 99.88 & 99.87 & 99.98 & 99.96 & 100.00 & 97.28 & 99.87 & 76.00 & 99.78 & 99.93 & 92.23 & 99.98 & 99.79 & 99.98 & 99.30 & 99.32 & 99.38 & 99.94 & 98.02 \\
\hline
\rowcolor{gray!20} 
FTNet (\textbf{ours}) & 4-shot 
& 100.0 & 99.91 & 99.13 & 100.0 & 99.90 & 100.0 & 92.32 & 98.31 & 87.83 & 100.0 & 99.97 & 99.80 & 100.0 & 100.0 & 100.0 & 100.0 & 100.0 & 100.0 & 100.0 & \underline{98.81} \\
\rowcolor{gray!20} 
\rowcolor{gray!20} 
FTNet-T (\textbf{ours}) & 4-shot 
& 100.0& 100.0 & 100.0 & 100.0 & 99.93 & 100.00 & 96.59 & 100.0 & 87.73 & 100.0 & 100.0 & 100.0 & 100.0 & 100.0 & 100.0 & 100.0 & 100.0 & 100.0 & 100.0 & \textbf{99.20} \\
\bottomrule
\end{tabular}%
} 
\caption{Comparison on the UniversalFakeDetect. The baseline results are directly cited from \cite{tan2025c2p}. All baseline methods are trained on the ProGAN dataset. \textbf{Bold} and \underline{underline} indicate the best and second-best performances, respectively.}
\label{tab:cross4}
\end{table*}

\renewcommand{\arraystretch}{1.2}
\setlength{\tabcolsep}{6pt}
\begin{table*}[t] 
\centering
\small
\begin{tabular}{l|c|cccccccc|c}
\toprule
Training Dataset & Number& CNNSpot &GramNet &LNP  & UnivFD & PatchCraft &DIRE &NPR & AIDE & FTNet(1-shot) \\
\midrule
ProGAN & 360K& 56.94&58.94 & 57.11& 57.22 & 53.76 & 58.19 &57.29 & 58.37 & 80.71(+\textbf{21.77\%}) \\
SD v1.4&324k & 60.11 &60.95 &55.63 & 55.62 & 56.32 &59.71 &58.13 & 62.60 & 80.71(+\textbf{18.11\%}) \\
All GenImage &2916K & 60.89 &59.81 &58.52  & 60.42 & 55.70 &57.83 &57.81 & 65.77 & 80.71(+\textbf{14.94\%}) \\
\bottomrule
\end{tabular}
\footnotetext{1-shot setting for FTNet; mAcc averaged over Human, Animal, Object, Scene categories.}
\caption{Comparison on the Chameleon. The baseline methods results are directly cited from paper\cite{yan2024sanity}. }
\label{tab:cross5}
\end{table*}

\noindent\textbf{ Evaluation on Chameleon} 
To test the generalization capabilities of our method under the most demanding conditions, we compared our performance against several state-of-the-art detectors on the challenging Chameleon dataset. The results are shown in Table \ref{tab:cross5}. The Chameleon dataset is designed to simulate extreme generalization scenarios, and its generative architecture differs significantly from the conventional training set. These experimental results reveal a common dilemma faced by existing methods: despite being trained on millions of images, these baseline methods almost universally fail on this dataset. Even the top-performing method, AIDE, achieves only 65.77\% accuracy. In stark contrast, our method achieves an impressive 80.71\% performance¹, requiring only a single sample (1-shot) for adaptation, an improvement of 14.94\% over the state-of-the-art AIDE, without requiring any parameter updates. This result once again highlights the fragility of the traditional zero-shot generalization paradigm in uncharted domains and strongly demonstrates the significant superiority and practical value of our few-shot framework for addressing complex, real-world challenges.
\renewcommand{\arraystretch}{1.4}
\setlength{\tabcolsep}{4pt}
\begin{table*}[h!]
\small
\centering
\resizebox{\textwidth}{!}{%
\begin{tabular}{c|cccccccccccccccccccc}
\toprule
\multirow{2}{*}{Method} & \multicolumn{2}{c}{AttGAN} & \multicolumn{2}{c}{BEGAN} & \multicolumn{2}{c}{CramerGAN} & \multicolumn{2}{c}{InfoMaxGAN} & \multicolumn{2}{c}{MMDGAN} & \multicolumn{2}{c}{RelGAN} & \multicolumn{2}{c}{S3GAN} & \multicolumn{2}{c}{SNGAN} & \multicolumn{2}{c}{STGAN} & \multicolumn{2}{c}{Mean} \\
\cmidrule(lr){2-3} \cmidrule(lr){4-5} \cmidrule(lr){6-7} \cmidrule(lr){8-9} \cmidrule(lr){10-11} \cmidrule(lr){12-13} \cmidrule(lr){14-15} \cmidrule(lr){16-17} \cmidrule(lr){18-19} \cmidrule(lr){20-21}
 & Acc. & A.P. & Acc. & A.P. & Acc. & A.P. & Acc. & A.P. & Acc. & A.P. & Acc. & A.P. & Acc. & A.P. & Acc. & A.P. & Acc. & A.P. & Acc. & A.P. \\
\midrule
CNNdet\cite{wang2020cnn}   & 51.1 & 83.7 & 50.2 & 44.9 & 81.5 & 97.5 & 71.1 & 94.7 & 72.9 & 94.4 & 53.3 & 82.1 & 55.2 & 66.1 & 62.7 & 90.4 & 63.0 & 92.7 & 62.3 & 82.9 \\
Frank\cite{frank2020leveraging}   & 65.0 & 74.4 & 39.4 & 39.9 & 31.0 & 36.0 & 41.1 & 41.0 & 38.4 & 40.5 & 69.2 & 96.2 & 69.7 & 81.9 & 48.4 & 47.9 & 25.4 & 34.0 & 47.5 & 54.7 \\
Durall\cite{durall2020watch}   & 39.9 & 38.2 & 48.2 & 30.9 & 60.9 & 67.2 & 50.1 & 51.7 & 59.5 & 65.5 & 80.0 & 88.2 & 87.3 & 97.0 & 54.8 & 58.9 & 62.1 & 72.5 & 60.3 & 63.3 \\
Patchfor\cite{chai2020makes} & 68.0 & 92.9 & 97.1 & 100.0 & 97.8 & 99.9 & 93.6 & 98.2 & 97.9 & 100.0 & 99.6 & 100.0 & 66.8 & 68.1 & 97.6 & 99.8 & 92.7 & 99.8 & 90.1 & 95.4 \\
F3Net\cite{qian2020thinking}   & 85.2 & 94.8 & 87.1 & 97.5 & 89.5 & 99.8 & 67.1 & 83.1 & 73.7 & 99.6 & 98.8 & 100.0 & 65.4 & 70.0 & 51.6 & 93.6 & 60.3 & 99.9 & 75.4 & 93.1 \\
SelfBland\cite{shiohara2022detecting} & 63.1 & 66.1 & 56.4 & 59.0 & 75.1 & 82.4 & 79.0 & 82.5 & 68.6 & 74.0 & 73.6 & 77.8 & 53.2 & 53.9 & 61.6 & 65.0 & 61.2 & 66.7 & 65.8 & 69.7 \\
LGrad\cite{tan2023learning}   & 68.6 & 93.8 & 69.9 & 89.2 & 50.3 & 54.0 & 71.1 & 82.0 & 57.5 & 67.3 & 89.1 & 99.1 & 78.5 & 86.0 & 78.0 & 87.4 & 54.8 & 68.0 & 68.6 & 80.8 \\
Ojha\cite{ojha2023towards}     & 78.5 & 98.3 & 72.0 & 98.9 & 77.6 & 99.8 & 77.6 & 98.9 & 77.6 & 99.7 & 78.2 & 98.7 & 85.2 & 98.1 & 77.6 & 98.7 & 74.2 & 97.8 & 77.6 & 98.8 \\
NPR\cite{tan2024rethinking}     & 83.0 & 96.2 & 99.0 & 99.8 & 98.7 & 99.0 & 94.5 & 98.3 & 98.6 & 99.0 & 99.6 & 100.0 & 79.0 & 80.0 & 88.8 & 97.4 & 98.0 & 100.0 & 93.2 & 96.6 \\
SAGNet\cite{tao2025sagnet} & 95.2 & 99.6 & 95.0 & 99.8 & 94.4 & 99.8 & 94.4 & 99.8 & 94.4 & 99.8 & 95.7 & 99.7 & 96.5 & 99.6 & 94.3 & 99.7 & 86.3 & 97.1 & \underline{94.0} & \underline{99.4} \\
\hline
FTNet(2-shot)& 100.0 & 100.0 & 100.0 & 100.0 & 96.8 & 100.0 & 96.8 & 100.0 & 96.8 & 100.0 & 99.9 & 100.0 & 80.2 & 99.6 & 96.87 & 100.0 & 99.9 & 100.0 & \textbf{96.4} & \textbf{100.0} \\
\bottomrule
\end{tabular}
}
\caption{Cross-GAN-Sources Evaluation on the Self-Synthesis 9 GANs dataset. \textbf{Bold} and \underline{underline} indicate the best and second-best performances, respectively.} 
\label{tab:GAN} 
\end{table*}

\noindent\textbf{Evaluation on GANGen} 
We show the results on the
 GANGen dataset in Table \ref{tab:GAN}. 
We only need two samples to achieve SOTA performance. Compared to the SAGNet, we achieve an improvement of 2.4\%. Perhaps this is due to the similarity of the generative architecture between GAN models, which makes CLIP able to distinguish different artifacts very well. However, since our sample size is very small, we believe this is our advantage. It is feasible to achieve high accuracy with a very small number of samples.
To clearly compare the performance of our method with existing baseline models, Figure \ref{fig-leida} uses a radar chart to show the evaluation results on multiple key indicators. As can be seen from the figure, the coverage area of our method is significantly better than that of other comparison methods.

\begin{table}[htbp]  
\centering
\small  
\setlength{\tabcolsep}{5pt}  
\renewcommand{\arraystretch}{1.2}  
\begin{tabular}{llccc}  
\toprule
Method & Backbone & Params & GenImage & Chameleon \\
\midrule
FTNet & VIT-L/14 & 304M & 90.7 &80.7 \\
FTNet & VIT-B/16 & 86M & 60.4 & 77.4\\
FTNet & VIT-B/32 & 86M  & 61.2 &77.1\\
\bottomrule
\end{tabular}

\caption{Performance Comparison on different backbones} 
\label{tab:genimage-performance}   
\end{table}


\subsection{Additional Ablation Studies}
\noindent\textbf{Effect of network backbones}
 We selected three different ViT (Vision Transformer) backbone networks: VIT-L/14, VIT-B/16, and VIT-B/32.The results are shown in the table\ref{tab:genimage-performance}. When the number of backbone model parameters is larger, the model's feature extraction ability is better and the classification ability is stronger. Selecting VIT-L/14 as the backbone, the average accuracy on three datasets is 86.3\%.
 
\begin{figure}[!htbp]
  \centering
   \includegraphics[width=\columnwidth]{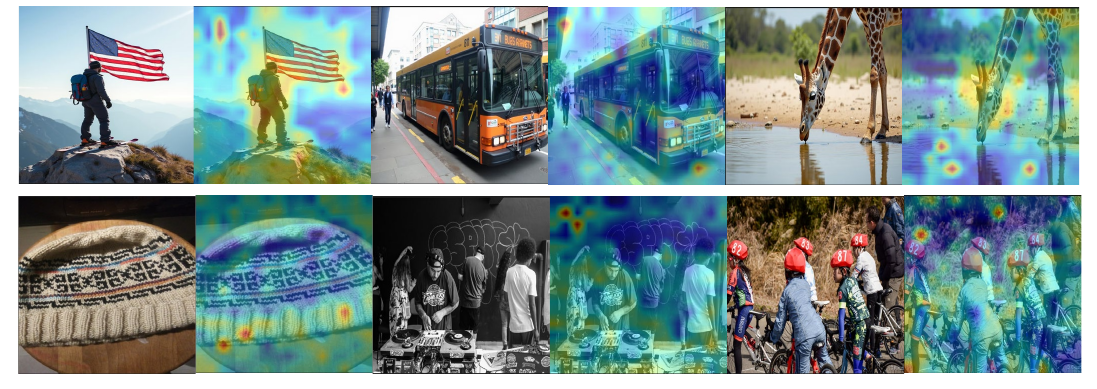}
   \caption{The visualization of CAM, the top and bottom rows display fake and real images}
   \label{fig:fig-8}
\end{figure}

 \noindent\textbf{Visualization of CAM}
To gain a deeper understanding of how CLIP distinguishes forged images, we analyzed CLIP layer 12 using class activation map (CAM) visualization \cite{zhou2016learning}. Figure \ref{fig:fig-8} shows the class activation maps for images from Flux.1. The upper and lower columns correspond to forged and real images, respectively. Notably, the CAMs for real images highlight broader regions of the image, while the CAMs for forged images tend to emphasize local regions. This suggests that CLIP's intermediate layers can detect subtle artifacts between forged images. Our approach fully exploits the potential of CLIP, achieving good generalization performance without requiring extensive training resources.

\begin{figure*}[!t]
\centering
\includegraphics[width=2.0\columnwidth, height=10cm, keepaspectratio]{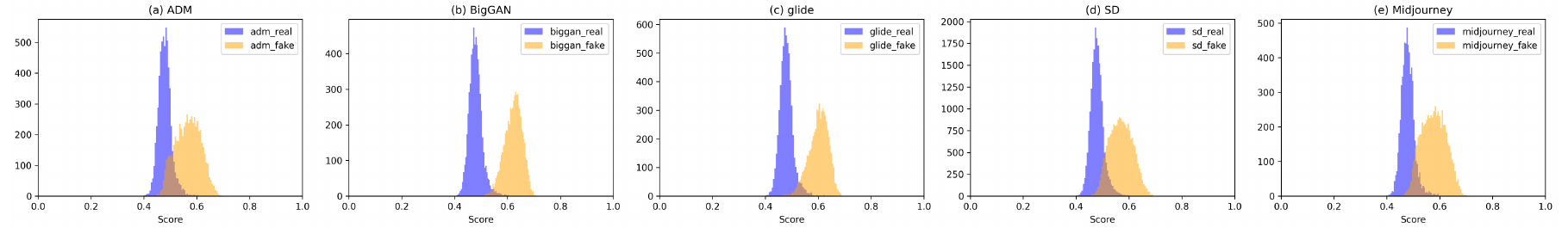}
\caption{Logit distributions of extracted forgery features. We tested on datasets such as ADM, BigGAN, glide, and SD.}
\label{fig-logits}
\end{figure*}

 \noindent\textbf{Logit Distributions Visualization}
To clearly demonstrate the advantages of our method, we visualize the Logit distribution of our method. As shown in the Figure \ref{fig-logits}. It can be seen that the logits distribution clearly distinguishes true and false, with little overlap.

\begin{figure}[!htbp]

  \centering
   \includegraphics[width=\columnwidth]{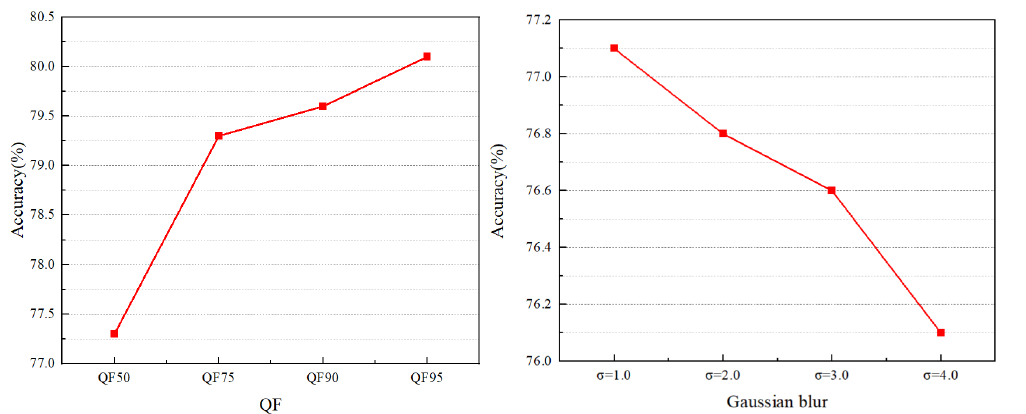}
   \vspace{-0.1in}
   \caption{Robustness on JPEG compression and Gaussian blur of FTNet about the Chameleon.}
   \vspace{-0.1in}
   \label{fig:fig-rob}
   
\end{figure}

 \noindent\textbf{Robustness Evaluation}
 
Images in the real world are often distorted by operations such as JPEG compression and Gaussian blur, making the detection of AI-generated images more challenging. This paper aims to evaluate the robustness of different detection methods to these perturbations. The perturbations tested include JPEG compression with different quality factors (QF=95, 90, 75, 50) and Gaussian blur with different standard deviations ($\sigma$=1.0, 2.0, 3.0, 4.0). We conduct experimental verification in Chameleon. As shown in the Figure \ref{fig:fig-rob}. Since our method is based on similarity for classification, it is greatly affected by compression or blurring, which is another direction we will focus on for optimization in the future. However, on the Chameleon dataset, our performance is still better than the baseline method.

 \noindent\textbf{Limitations and Future Work}
 While the proposed method performs well in low-sample scenarios, its robustness is inferior to that of models trained on a larger scale. It has limited ability to detect locally manipulated images and exhibits poor robustness to common image perturbations such as compression and Gaussian blur. Furthermore, the high sample quality requirements pose a challenge. Therefore, future work could focus on enhancing the ability to detect localized forgeries, improving model robustness, and exploring more effective adaptive strategies to overcome performance bottlenecks.

\newpage

\bibliography{aaai2026}

\end{document}